%% file: main.tex
\documentclass[10pt,twocolumn,letterpaper]{article}

\usepackage{iccv}              % To produce the CAMERA-READY version

\definecolor{iccvblue}{rgb}{0.21,0.49,0.74}
\usepackage[pagebackref,breaklinks,colorlinks,allcolors=iccvblue]{hyperref}
\usepackage{bm}
\usepackage{makecell}
\usepackage{url}

\usepackage{arydshln}

\def\paperID{19} % *** Enter the Paper ID here
\def\confName{ICCV}
\def\confYear{2025}

\title{Efficient Face Image Quality Assessment via Self-training and Knowledge Distillation}

\author{Wei Sun$^{1}$, Weixia Zhang$^{2}$, Linhan Cao$^{2}$, Jun Jia$^{2}$, \\ Xiangyang Zhu$^{3}$, Dandan Zhu$^{1}$\thanks{Corresponding authors.}, Xiongkuo Min$^{2}$, Guangtao Zhai$^{2}$\footnotemark[1]\\
$^1$East China Normal University, $^2$Shanghai Jiao Tong University, \\$^3$Shanghai Artificial Intelligence Laboratory
}
\begin{document}
\maketitle
\begin{abstract}
Face image quality assessment (FIQA) is essential for various face-related applications. Although FIQA has been extensively studied and achieved significant progress, the computational complexity of FIQA algorithms remains a key concern for ensuring scalability and practical deployment in real-world systems. In this paper, we aim to develop a computationally efficient FIQA method that can be easily deployed in real-world applications. Specifically, our method consists of two stages: training a powerful teacher model and distilling a lightweight student model from it. To build a strong teacher model, we adopt a self-training strategy to improve its capacity. We first train the teacher model using labeled face images, then use it to generate pseudo-labels for a set of unlabeled images. These pseudo-labeled samples are used in two ways: (1) to distill knowledge into the student model, and (2) to combine with the original labeled images to further enhance the teacher model through self-training. The enhanced teacher model is used to further pseudo-label another set of unlabeled images for distilling the student models. The student model is trained using a combination of labeled images, pseudo-labeled images from the original teacher model, and pseudo-labeled images from the enhanced teacher model. Experimental results demonstrate that our student model achieves comparable performance to the teacher model with an extremely low computational overhead. Moreover, our method achieved first place in the ICCV 2025 VQualA FIQA Challenge. The code is available at \url{https://github.com/sunwei925/Efficient-FIQA.git}.

\end{abstract}
\section{Introduction}
\label{sec:intro}

Face image quality assessment (FIQA)~\cite{schlett2022face} plays a vital role in various computer vision applications, particularly those involving face-related tasks. For example, FIQA can help optimize face image enhancement algorithms, filter out low-quality facial images to improve the robustness and reliability of face recognition systems, and support more effective recommendation strategies for facial images on social media platforms.

Existing FIQA studies can be broadly categorized into two types. The first is \textit{perceptual FIQA}~\cite{su2023going,liu2024assessing,chen2024dsl}, which aims to assess the visual quality of facial images based on human subjective perception. The second is \textit{recognition-oriented FIQA}~\cite{hernandez2019faceqnet,terhorst2020ser,ou2021sdd}, which focuses on evaluating the recognition quality of facial images based on their effectiveness in face recognition systems. In this paper, we focus on perceptual FIQA, which has broader applications than recognition-oriented FIQA and is more aligned with human subjective perception.

For perceptual FIQA, the typical research pipeline involves collecting a large-scale facial image dataset and conducting subjective experiments to obtain mean opinion scores (MOSs) from human subjects. Popular FIQA datasets, such as GFIQA-20K~\cite{su2023going} and CGFIQA-40K~\cite{chen2024dsl}, typically consist of over ten thousand Internet-sourced facial images annotated with corresponding MOS labels. These labeled datasets are then used to train facial feature-aware deep neural networks (DNNs), which leverage priors such as generative face models~\cite{su2023going}, facial landmarks~\cite{chen2024dsl}, and other semantic features to achieve high performance in face image quality assessment.

However, the majority of these models are computationally intensive, requiring large-scale neural network parameters and substantial computational resources for inference. Such complexity poses challenges for deploying FIQA models in real-world scenarios, especially in resource-constrained environments such as mobile devices, surveillance systems, or large-scale commercial platforms that demand real-time processing. To address these challenges, we propose a simple yet effective training framework for efficient FIQA, built upon iterative self-training and knowledge distillation. 

Our framework consists of two main stages. In the first stage, we construct a strong teacher model using a self-training paradigm. Specifically, the teacher is initially trained on a labeled FIQA dataset. Once trained, it is used to generate pseudo-labels for a set of unlabeled facial images. These pseudo-labeled samples serve two key roles: (1) they act as soft supervision for training a lightweight student model, and (2) they are combined with the labeled data to further fine-tune the teacher, thereby progressively enhancing its predictive capacity through iterative self-training. In the second stage, the enhanced teacher---now with improved generalization---is used to generate additional pseudo-labels on a new set of unlabeled data. The student model is then trained using a comprehensive supervision set, comprising (1) the original labeled data, (2) pseudo-labeled data from the initial teacher, and (3) pseudo-labeled data from the enhanced teacher. This staged supervision strategy enables the student model to inherit both the original and refined knowledge from the teacher. Experimental results demonstrate that the distilled student model achieves performance comparable to the teacher model, while significantly reducing computational cost. The student model contains only 1 million parameters and requires 0.4949 GFLOPs, resulting in a 99\% reduction in parameters and nearly 100x lower FLOPs compared to the teacher. Moreover, our method achieved first place in the FIQA Challenge~\cite{ma2025fiqa} held by ICCV VQualA 2025~\cite{isrgcq2025iccvw,zhu2025vqa,genai-bench2025iccvw,li2025evqa,ma2025fiqa,diqa2025iccvw}.

The main contributions of this work are summarized as follows:
\begin{itemize}
\item We propose a two-stage knowledge distillation framework for FIQA, which comprises a self-trained teacher model and a distilled lightweight student model.
\item We introduce an iterative pseudo-labeling strategy that progressively enhances the teacher model and improves the quality of supervision for student training.
\item We demonstrate that the distilled student model achieves performance comparable to the teacher model, while significantly reducing computational complexity.
\end{itemize}

\section{Related Work}
\label{sec:related_work}

\subsection{Image Quality Assessment}

Image quality assessment (IQA)~\cite{zhai2020perceptual} methods can be broadly classified into two categories: handcrafted feature-based methods and DNN-based methods. Handcrafted feature-based methods usually rely on natural scene statistics (NSS)~\cite{ruderman1994statistics,mittal2012making,mittal2012no,moorthy2011blind} or other manually designed features such as texture~\cite{narvekar2011no,sebastian2012gray}, edge~\cite{ferzli2009no,bahrami2014fast,vu2009s3}, or free energy cues~\cite{zhai2011psychovisual}. For example, BRISQUE~\cite{mittal2012no} models natural scene statistics of locally normalized luminance coefficients and their pairwise products in the spatial domain to quantify perceptual quality. NIQE~\cite{mittal2012making} quantifies perceptual quality by measuring deviations from statistical regularities in locally normalized luminance features of natural image patches, without requiring training on human opinion scores. The handcrafted feature-based methods~\cite{mittal2012making,mittal2012no,moorthy2011blind,zhang2021no,zhang2022no,zhai2021perceptual} are primarily designed for synthetically distorted images, but often struggle to generalize to real-world (in-the-wild) scenarios due to their limited representational capacity. 

In contrast, DNN-based IQA methods capitalize on the powerful representation learning capabilities of DNNs to automatically extract quality-aware features, thereby achieving superior performance across both synthetically distorted and authentically degraded images. Early DNN-based IQA models, such as CNNIQA\cite{kang2014convolutional} and WaDIQaM~\cite{bosse2017deep} utilize relatively shallow convolutional neural networks (CNNs) that operate on local image patches, and relied on simple pooling strategies, such as average pooling or weighted average pooling, to produce image-level quality predictions. With the rapid development of DNNs, recent IQA works~\cite{ke2021musiq,sun2023blind,chen2024topiq,wang2023exploring,yang2022maniqa,wu2024q,zhang2023blind,sun2024assessing,lu2023bh,yi2021attention,sun2021deep,sun2025compressedvqa,zhang2023md,zhang2024benchmarking,zhang2024human} have embraced more advanced and expressive network architectures. These include multi-scale feature extractors~\cite{ke2021musiq,sun2023blind,sun2022deep,sun2019mc360iqa} that capture visual features across different layers of DNNs, and attention mechanisms~\cite{yang2022maniqa,chen2024topiq,yi2021attention,cao2023attention} that selectively emphasize distortion-relevant regions, thereby improving perceptual sensitivity. In addition, the integration of high-level semantic priors from CLIP variants~\cite{wang2023exploring,zhang2023blind,zhang2025advancing}, as well as the fine-tuning of large multimodal models (LMMs)~\cite{wu2024q,ge2025lmm,wu2025fvq,li2025aghi,cao2025agav,wang2024large}, opens new avenues for quality assessment by jointly leveraging visual and textual modalities to capture more comprehensive quality semantics. To enhance cross-dataset robustness and real-world applicability, several studies have explored generalization-oriented strategies such as multi-dataset training~\cite{zhang2023blind,sun2023blind}, which helps mitigate dataset-specific biases. In parallel, unsupervised and self-supervised learning paradigms~\cite{madhusudana2022image,agnolucci2024arniqa,cao2025breaking} have gained attention for leveraging large-scale unlabeled data, enabling models to learn robust quality representations without relying on extensive human-annotated labels.

\subsection{Face Image Quality Assessment}

Perceptual FIQA aims to evaluate the visual quality of facial images as perceived by human observers, which is crucial for both human-centered media applications. To support research in this area, a series of FIQA datasets and models have been proposed. Su \textit{et al.}~\cite{su2023going}  introduce GFIQA-20K, a large-scale FIQA dataset consisting of 20,000 facial images collected from the Internet, each annotated with subjective quality scores obtained through human evaluation. They also proposed a generative prior-guided IQA model that integrates generative facial priors into the quality prediction process, enabling the model to better capture high-level facial semantics and perceptual degradation. Chahine \textit{et al.}~\cite{chahine2023image} present a portrait IQA dataset containing 5,116 high-resolution images captured in 50 diverse real-world scenes. The dataset includes expert-labeled pairwise comparisons across multiple perceptual dimensions such as detail, exposure, and overall quality. To model these fine-grained attributes, they proposed SEM-HyperIQA, a semantics-aware architecture that adapts to scene-specific quality scales by incorporating both local and contextual features. Sun \textit{et al.}~\cite{sun2024dual} propose a dual-branch network for portrait IQA that employs separate Swin Transformer backbones to extract features from the full image and the facial region, and is optimized using a learning-to-rank strategy. Liu \textit{et al.}~\cite{liu2024assessing} address the challenges posed by both authentic and synthetic distortions in facial images. They constructed a hybrid FIQA dataset incorporating both in-the-wild and artificially degraded images. They then propose TransFQA, a two-branch Transformer-based model, that jointly captures global context, facial structure, and facial component-level cues while integrating distortion-aware features to improve the robustness and accuracy of quality predictions across diverse conditions. To further address fairness and generalization, Chen \textit{et al.}~\cite{chen2024dsl} propose CGFIQA-40K, a comprehensive and demographically balanced dataset containing 40,000 facial images, with careful control over gender and skin tone distributions. They introduce DSL-FIQA, a dual-set learning framework that incorporates both real-world and synthetic distortions via a degradation representation learning strategy and integrates facial landmark-guided attention mechanisms to focus on perceptually important regions such as eyes, mouth, and skin texture, thereby enhancing interpretability and alignment with human perception.

\subsection{Self-training and Knowledge Distillation}

\textbf{Self-training} begins by training a reliable teacher model on labeled data, which is then used to generate pseudo-labels for unlabeled data. The original labeled data and the newly pseudo-labeled data are subsequently combined to train a more robust student model. Typically, self-training incorporates noise into the training samples or the model itself to increase the learning difficulty, thereby encouraging the model to learn more powerful and generalizable feature representations. For example, Yalniz \textit{et al.}~\cite{yalniz2019billion} demonstrate a large-scale self-training framework for image classification, where a teacher model trained on labeled data is used to pseudo-label up to one billion unlabeled images. A student model is then trained on this pseudo-labeled dataset and fine-tuned on clean labeled data, resulting in substantial performance gains and showing the effectiveness of self-training at scale. Similarly, Xie \textit{et al.}~\cite{xie2020self} propose Noisy Student Training, a self-training method that improves ImageNet classification by iteratively training a student model on both labeled and pseudo-labeled data. The student, which is equal to or larger than the teacher, is trained with added noise, such as dropout and data augmentation, to improve generalization. However, since this work focuses on assessing the visual quality of facial images, we avoid introducing noise to the input data during self-training, as it may compromise the perceptual fidelity of the images.

\textbf{Knowledge distillation} is a model compression technique in which a high-capacity teacher model transfers its knowledge to a smaller student model. Hinton \textit{et al.}~\cite{hinton2015distilling} first introduce the teacher-student learning framework, where the student is trained on softened class probabilities (soft targets) instead of hard labels, enabling it to inherit the generalization ability of the teacher with reduced complexity. Furlanello \textit{et al.}~\cite{furlanello2018born} propose born-again Networks, where student models with the same architecture as their teachers are trained through knowledge distillation. Interestingly, these students often outperform their teachers, demonstrating the potential of self-distillation even without model compression. For regression tasks such as image quality assessment, prior studies have shown that using pseudo-labels generated by a teacher model on large-scale unlabeled data can further improve distillation performance. For instance, Sun \textit{et al.}~\cite{sun2025empirical} proposed E-VQA, an efficient video quality assessment model that leverages knowledge distillation. A high-capacity teacher model, RQ-VQA~\cite{sun2024enhancing}, generates pseudo-labels on a large unlabeled video dataset, which are then used to pre-train E-VQA. This approach significantly enhances model accuracy while maintaining low computational cost, highlighting the effectiveness of knowledge distillation for quality assessment. In this work, we adopt a similar knowledge distillation strategy inspired by E-VQA.

\section{Proposed Method}
\begin{figure}[t]
    \centering
    \includegraphics[width=0.9\linewidth]{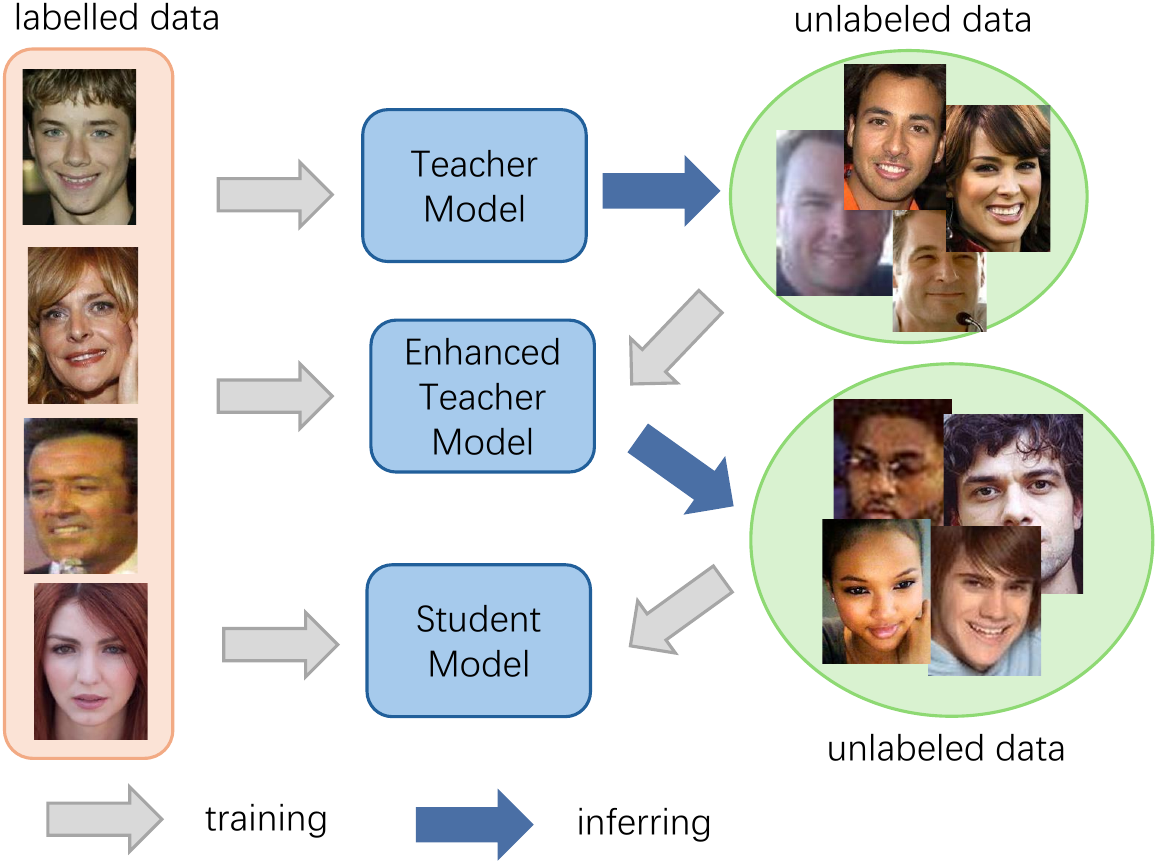}
    \caption{The overall architecture of our proposed FIQA framework. We begin by training a teacher model on a labeled dataset. The trained teacher model is then used to generate pseudo labels for unlabeled facial images. These pseudo-labeled samples are used both to refine the teacher model through self-training and to supervise the distillation process. The the enhanced teacher model further generates pseudo labels from an additional unlabeled set. Finally, a student model is trained using a combination of labeled data and pseudo-labeled data from both the original and enhanced teacher models. Grey arrows denote training processes, while blue arrows represent inference and pseudo-label generation.}
    \label{fig:framework}
\end{figure}
As illustrated in Figure~\ref{fig:framework}, our method can be divided into two stages: self-training for the teacher model and knowledge distillation for the student model.

\subsection{Teacher Model Optimization via Self-Training}
We first aim to train a strong teacher model using a self-training strategy that leverages both accurately labeled data and the generalization potential of unlabeled data to enhance the teacher model’s performance. Specifically, given a labeled training set $\mathcal{D} = \{ (x_i, y_i) \}_{i=1}^{n}$, we first learn the teacher model $\theta^t_*$ by minimizing the loss function over the labeled data:

\begin{equation}
\frac{1}{n} \sum_{i=1}^{n} \mathcal{L} \left(y_i, f^t(x_i, \theta^t)\right),
\end{equation}
where $f^t$ denotes the teacher network.

Next, we use the trained teacher model $f^t(\theta^t_*)$ to generate pseudo labels for a set of unlabeled images:
\begin{equation}
\tilde{y}_i = f(\tilde{x}_i, \theta_*^t), \quad \forall i = 1, \cdots, m,
\end{equation}
where $\tilde{x}_i$ denotes an unlabeled input sample, and $\tilde{y}_i$ is the corresponding pseudo-label generated by the trained teacher model.

The pseudo-labeled data $\mathcal{D}_{\text{pseudo}}^{(1)} = \{ (\tilde{x}_i, \tilde{y}_i) \}_{i=1}^{m}$ serves two purposes. First, it is used as soft supervision to distill a lightweight student model from the trained teacher model in the second stage. Second, we leverage both the pseudo-labeled data $\mathcal{D}_{\text{pseudo}}^{(1)}$ and the original labeled data $\mathcal{D}$ to learn the teacher model $\theta^{t+}_*$, further enhancing its accuracy and generalization ability:
\begin{equation}
\frac{1}{n} \sum_{i=1}^{n} \mathcal{L}(y_i, f^t(x_i, \theta^{t+}) + \frac{1}{m} \sum_{i=1}^{m} \mathcal{L}(\tilde{y}_i, f^t(\tilde{x}_i, \theta^{t+}).
\end{equation}
% where $\hat{\theta}^t$ denotes the parameters of the enhanced teacher model.

To distill knowledge from the enhanced teacher, we further use the model $\theta^{t+}_*$ to generate pseudo-labels for an additional set of unlabeled images, resulting in another pseudo-labeled dataset $\mathcal{D}_{\text{pseudo}}^{(2)} = \{ (\tilde{x}_i, \tilde{y}_i) \}_{i=1}^{m^\prime}$.

\subsection{Student Model Optimization via Knowledge Distillation}
After obtaining strong teacher models through self-training, we distill a lightweight student model from the teacher models. The goal is to transfer the knowledge embedded in the high-capacity teacher models to a compact student model with significantly lower computational complexity, while preserving performance.

To achieve this, we construct a comprehensive supervision set consisting of three sources: (1) the original labeled dataset $\mathcal{D}$, (2) the pseudo-labeled dataset $\mathcal{D}_{\text{pseudo}}^{(1)}$ generated by the initial teacher model, and (3) the additional pseudo-labeled dataset $\mathcal{D}_{\text{pseudo}}^{(2)}$ generated by the enhanced teacher model:

\begin{align}
\mathcal{L}_{s} = \
& \frac{1}{n} \sum_{i=1}^{n} \mathcal{L}(y_i, f^{s}(x_i, \theta^s)) \nonumber \\
+ & \frac{1}{m} \sum_{i=1}^{m} \mathcal{L}(\tilde{y}_i, f^{s}(\tilde{x}_i, \theta^s)) \nonumber \\
+ & \frac{1}{m'} \sum_{i=1}^{m'} \mathcal{L}(\bar{y}_i, f^{s}(\bar{x}_i, \theta^s)), \label{eq:student_loss}
\end{align}
where $\theta^s$ and $f^{s}$ denotes the parameters and network structure of the student model.

\subsection{Loss Function}
To supervise the training of both teacher and student models, we adopt two complementary regression loss functions: Mean Squared Error (MSE) and Pearson Linear Correlation Coefficient (PLCC) loss. MSE focuses on minimizing the absolute error between the predicted and ground-truth scores while PLCC aims to maximize the linear correlation between them. The combination of both losses ensures that the model predictions are not only numerically accurate but also rank-consistent with human ratings. 

The MSE loss is defined as:

\begin{equation}
\mathcal{L}_{\text{MSE}} = \frac{1}{N} \sum_{i=1}^{N} (y_i - \hat{y}_i)^2
\end{equation}
where $y_i$ denotes the ground-truth MOS for the $i-$-th sample, and $\hat{y}_i$ is the predicted quality score.

The PLCC loss is formulated as the negative Pearson correlation between predicted and ground-truth scores:

\begin{equation}
\begin{aligned}
\mathcal{L}_{\text{PLCC}} &= (1-\frac{\langle \bm{\hat{y}} - {\rm mean}(\bm{\hat{y}}), \bm {y} - {\rm mean}(\bm{y}) \rangle}{\Vert \bm{\hat{y}} - {\rm mean}(\bm{\hat{y}}) \Vert_2 \Vert {\bm y} - {\rm mean}(\bm{y}) \Vert_2})/2, \\
\end{aligned}
\end{equation}
where ${\bm y}$ and $\bm {\hat{y}}$ are the vectors of ground-truth and predicted quality scores of the face images in a batch respectively, $ {\langle \cdot \rangle}$ represents the inner product of two vectors, $\Vert\cdot\Vert$ denotes the norm operator for a vector, and $\rm mean$ is the average operator for a vector.

The total loss used to train the models is a weighted combination of the two:

\begin{equation}
\mathcal{L}_{\text{total}} = \mathcal{L}_{\text{MSE}} + \lambda \cdot \mathcal{L}_{\text{PLCC}},
\end{equation}
where $\lambda$ is a hyperparameter that balances the contribution of the two losses. In our experiments, $\lambda$ is set to 1.

\section{Experiments}
\subsection{Datasets}
We utilize the released training dataset from the ICCV 2025 VQualA FIQA Challenge as the labeled data to train our models. This dataset contains 27,686 face images collected from CelebA and Flickr. We randomly split the training data into 80\% for training and 20\% as a private validation set for selecting the best-performing model. We evaluate the performance of the trained models on both the validation and test sets of the challenge, which contain 1,000 and 889 images, respectively. The validation set includes samples from both CelebA and Flickr, while the test set consists exclusively of Flickr images.

For the unlabeled data, we collect approximately 400,000 face images from the Internet. Two subsets of 200,000 images each are used for pseudo-labeling in the first and second stages of our training framework, respectively.

\subsection{Experiment Setups}
We adopt the Swin Transformer Base (Swin-B)~\cite{liu2021swin} as the backbone for the teacher model. The original classification head is removed and replaced with a two-layer multilayer perceptron (MLP) consisting of 128 and 1 hidden neurons, respectively, serving as a regression head for quality prediction. For the student model, we use EdgeNeXt-XX-Small~\cite{maaz2022edgenext} as the backbone, similarly replacing its classification head with the same two-layer MLP.

The entire framework is implemented using PyTorch 2.4. For the two-round teacher training, we use the AdamW optimizer with a learning rate of $1 \times 10^{-4}$, weight decay of $1 \times 10^{-6}$, and a learning rate decay factor of 0.1 every 10 epochs. The model is trained for 30 epochs with a batch size of 32. We randomly select 20\% of the provided labeled data as a validation set to determine the best-performing teacher model. During training, images are resized such that the small side is 448 pixels while preserving the aspect ratio, followed by a random crop of $448 \times 448$. For student training, the only difference is the image resolution: images are resized and cropped to $352 \times 352$ instead. The teacher model is trained using two NVIDIA RTX 3090 GPUs, while the student model is trained on a single RTX 3090 GPU.

\subsection{Compared Methods}
We compared our method with six IQA approaches, including two handcrafted feature-based methods: NIQE~\cite{mittal2012making} and BRISQUE~\cite{moorthy2011blind}; three deep learning-based methods: MUSIQ~\cite{ke2021musiq}, StairIQA~\cite{sun2023blind}, and LIQE~\cite{zhang2023blind}; and a lightweight baseline implemented using MobileNetV2~\cite{sandler2018mobilenetv2}. Our method includes three variants: two teacher models and one student model. For fairness, all methods were retrained on the three test datasets, except NIQE, which is unsupervised and does not rely on human opinion scores.

\subsection{Evaluation Metrics}
We use Spearman’s Rank Correlation Coefficient (SRCC) and PLCC to evaluate the performance of the FIQA models, and report the number of parameters and GFLOPs to assess their efficiency.

\subsection{Experimental Results}
\begin{table}[t]
\small
\centering
\resizebox{0.48\textwidth}{!}{
\begin{tabular}{lcccc}
\toprule
\textbf{Method} & \textbf{SRCC} & \textbf{PLCC} & \textbf{Para. (M)} & \textbf{GFLOPs} \\
\midrule
NIQE~\cite{mittal2012making} & 0.5384 & 0.4765& ---&--- \\
BRISQUE~\cite{moorthy2011blind} & 0.8415 & 0.9145& ---&--- \\
MUSIQ~\cite{ke2021musiq} &  0.8953& 0.9070& 78.59 &230.74 \\
StairIQA~\cite{sun2023blind} & 0.9766 & 0.9852& 30.49&20.86 \\
LIQE~\cite{zhang2023blind} & 0.9655 & 0.9184& 151.28&3,013.76 \\
Teacher           & 0.9863 & 0.9924 & 86.88  & 50.13  \\
Teacher$+$  & 0.9867 & 0.9926 & 86.88  & 50.13  \\
\hdashline
MobileNetV2~\cite{sandler2018mobilenetv2}                & 0.9183 & 0.9522 & 3.251 & 0.3139 \\
Student           & 0.9855 & 0.9914 & 1.180   & 0.4949 \\
\bottomrule
\end{tabular}
}
\caption{Performance comparison of compared methods, teacher, enhanced teacher, and student models in terms of SRCC, PLCC, parameter count (in millions), and GFLOPs.}
\label{tab:fiqa_results}
\end{table}

Table~\ref{tab:fiqa_results} presents a comprehensive comparison of performance and efficiency among the compared methods, the teacher model, the enhanced teacher model (Teacher+), and the final student model. The handcrafted feature-based methods, NIQE and BRISQUE, perform poorly on the face image quality assessment task, suggesting that the NSS features they rely on are not effective for assessing face quality. In contrast, deep learning-based methods such as MUSIQ, StairIQA, and LIQE demonstrate stronger performance. Notably, our teacher model, despite having fewer parameters and lower computational complexity (GFLOPs) than models like MUSIQ and LIQE, achieves significantly better performance, validating the effectiveness of the selected architecture. To further enhance performance, we refine the teacher model using self-training with pseudo-labeled data. The enhanced teacher model (Teacher+) achieves a slight yet consistent gain over the original teacher, indicating that iterative self-labeling helps improve generalization. This progressively optimized teacher is then used to guide the training of a lightweight student model through knowledge distillation.

The lightweight baseline model, MobileNetV2, achieving moderate performance. Our student model, although much smaller in scale, achieves performance close to that of the teacher and enhanced teacher, with only 1.18 million parameters and 0.4949 GFLOPs. This represents a 99\% reduction in parameter count and nearly 100× lower computational cost compared to the teacher model. Despite this extreme efficiency, the student still maintains competitive accuracy, which highlights the strength of our distillation strategy in producing high-performing and deployable FIQA models.

\begin{table}[t]
\centering
\scriptsize
\begin{tabular}{ccc|cc}
\toprule
\makecell{\textbf{Labeled} \\ \textbf{Data}} & 
\makecell{\textbf{Pseudo Labels} \\ \textbf{(1st)}} & 
\makecell{\textbf{Pseudo Labels} \\ \textbf{(2nd)}} & 
\textbf{SRCC} & \textbf{PLCC} \\
\midrule
\checkmark &  &  & 0.9807 & 0.9884 \\
\checkmark & \checkmark &  & 0.9845 & 0.9907 \\
\checkmark & \checkmark & \checkmark & 0.9855 & 0.9914 \\
\bottomrule
\end{tabular}
\caption{Ablation study on the impact of using pseudo labels from different stages of the teacher model. All settings train the same student model.}
\label{tab:ablation_pseudo}
\end{table}

\subsection{Ablation Studies}

\begin{table}[t]
\small
\centering
\begin{tabular}{lcccc}
\toprule
\textbf{Resolution} & \textbf{SRCC} & \textbf{PLCC} & \textbf{Para. (M)} & \textbf{GFLOPs} \\
\midrule
$336\times336$                & 0.9848 & 0.9905 &1.180  & 0.4442 \\
$352\times352$           & 0.9855 & 0.9914 & 1.180   & 0.4949  \\
\bottomrule
\end{tabular}
\caption{Performance comparison of student models in terms of different resolutions.}
\label{tab:abliation_resolution}
\end{table}

Table~\ref{tab:ablation_pseudo} presents an ablation study on the effect of incorporating pseudo labels generated from different stages of the teacher model. When training the student model using only labeled data, the performance is reasonable but limited. Adding pseudo labels from the initial teacher significantly improves both SRCC and PLCC, indicating that unlabeled data---when guided by a well-trained teacher---can provide useful supervision. Further introducing pseudo labels generated by the enhanced teacher leads to the best performance, demonstrating that progressive refinement of the teacher model enhances the quality of pseudo labels and further benefits student learning. These results validate the effectiveness of our two-stage pseudo-labeling and distillation strategy.

Table~\ref{tab:abliation_resolution} compares the performance of student models trained with different input resolutions. Increasing the resolution from $336\times336$ to  $352\times352$ leads to a slight improvement in both SRCC and PLCC, while maintaining the same number of parameters. Although the higher resolution incurs a modest increase in GFLOPs, it results in better quality prediction, indicating that finer spatial details benefit FIQA performance.

\subsection{Performance on the FIQA Challenge}

\begin{table}[t]
\centering
\scriptsize
\begin{tabular}{clccc}
\toprule
\textbf{Rank} & \textbf{Team} & \textbf{Score} & \textbf{GFLOPs} & \textbf{Params (M)} \\
\midrule
1 & ECNU-SJTU VQA Team (Ours) & 0.9664 & 0.3313 & 1.1796 \\
2 & MediaForensics     & 0.9624 & 0.4687 & 1.5189 \\
3 & Next               & 0.9583 & 0.4533 & 1.2224 \\
4 & ATHENAFace         & 0.9566 & 0.4985 & 2.0916 \\
5 & NJUPT-IQA-Group    & 0.9547 & 0.4860 & 3.7171 \\
6 & ECNU VIS Lab       & 0.9406 & 0.4923 & 3.2805 \\
\bottomrule
\end{tabular}
\caption{Top-6 challenge results on the FIQA track, ranked by the official final score. The score is computed as the average of SRCC and PLCC.}
\label{tab:top6_results}
\end{table}

Table~\ref{tab:top6_results} shows the top-6 results of the ICCV 2025 VQualA FIQA Challenge. Our method, submitted under the name ECNU-SJTU VQA Team, ranked first with the highest overall score of 0.9664, calculated as the average of SRCC and PLCC. This result demonstrates the superior quality prediction capability of our approach.

In addition to its leading accuracy, our model maintains excellent efficiency. During inference, we adopt a resolution of $288\times288$, enabling the model to operate with only 0.3313 GFLOPs and 1.18 million parameters. Compared to other top-ranked teams, our method achieves the best trade-off between performance and computational cost, making it highly suitable for deployment in resource-constrained scenarios.

\section{Conclusion}
We proposed an efficient FIQA framework that integrates self-training and knowledge distillation to construct a lightweight yet effective face image quality assessment model. To improve the quality of supervision, we adopt a two-stage pseudo-labeling strategy, where a high-capacity teacher model is first trained on labeled data and then iteratively enhanced using pseudo-labeled samples generated from unlabeled data. The student model is distilled using a combination of labeled data and pseudo labels produced by both the original and enhanced teacher models. This progressive distillation process enables the student model to inherit strong quality prediction capabilities while maintaining a highly compact architecture. Extensive experiments demonstrate that our student model achieves performance comparable to the teacher model with significantly lower computational complexity. Our method ranked first in the ICCV 2025 VQualA FIQA Challenge, highlighting its effectiveness and potential for real-world deployment.

\section{Acknowledgement}
This work was supported in part by the National Natural Science Foundation of China under Grants 62301316, 62377011, and 62371283.
{
    \small
    \bibliographystyle{ieeenat_fullname}
    \bibliography{main}
}

\end{document}